\documentclass[10pt,twocolumn,letterpaper]{article}

\usepackage[pagenumbers]{wacv} 

\definecolor{wacvblue}{rgb}{0.21,0.49,0.74}
\usepackage[pagebackref,breaklinks,colorlinks,allcolors=wacvblue]{hyperref}

\def\approach{MultiID}
\def\benchmark{IDBench}

\usepackage{booktabs}
\usepackage{multirow}
\usepackage{amsmath}
\usepackage{colortbl}
\usepackage{pgfplots}
\usepackage{xcolor}
\usepackage{pgfmath}
\usepackage{circledsteps}

\definecolor{yzybest}{rgb}{0.96, 0.57, 0.58}
\definecolor{yzysecond}{rgb}{0.98, 0.78, 0.57}
\definecolor{yzythird}{rgb}{1.0, 1.0, 0.56}

\newcommand{\bestmark}[1]{\textbf{#1}}
\newcommand{\secondmark}[1]{\underline{#1}}

\definecolor{myred}{rgb}{1, 0.61, 0.66}
\definecolor{mygreen}{rgb}{0.60, 0.88, 1}

\newcommand{\colorize}[3]{
  \pgfmathsetmacro{\midpoint}{(#1 + #2)/2}%
  \pgfmathsetmacro{\scaled}{%
    ifthenelse(
      #3 < \midpoint,
      (#3 - #1)/(\midpoint - #1 + 0.0001), 
      (#3 - \midpoint)/(#2 - \midpoint + 0.0001) 
    )
  }%
  \pgfmathsetmacro{\colorvalue}{int(max(0, min(1, \scaled)) * 100)}%
  \pgfmathparse{#3 < \midpoint ? 1 : 0}
  \ifnum\pgfmathresult=1
    \edef\temp{\noexpand\cellcolor{white!\colorvalue!mygreen}}%
    \temp#3%
  \else
    \edef\temp{\noexpand\cellcolor{myred!\colorvalue!white}}%
    \temp#3%
  \fi
}

\def\wacvPaperID{2082} 
\def\confName{WACV}
\def\confYear{2026}

\title{A Training-Free Approach for Multi-ID Customization via Attention Adjustment and Spatial Control}

\author{
Jiawei Lin$^{\dagger}$\\
Xi'an Jiaotong University\\
\and
Guanlong Jiao$^{\dagger}$\\
Tsinghua University\\
\and
Jianjin Xu\\
Carnegie Mellon University\\
}

\begin{document}
\maketitle
\begingroup
\renewcommand\thefootnote{}\footnote{$^{\dagger}$These authors contributed equally to this work.}
\addtocounter{footnote}{-1}
\endgroup
\begin{abstract}

Multi-ID customization is an interesting topic in computer vision and attracts considerable attention recently.
Given the ID images of multiple individuals, its purpose is to generate a customized image that seamlessly integrates them while preserving their respective identities.
Compared to single-ID customization, multi-ID customization is much more difficult and poses two major challenges.
First, since the multi-ID customization model is trained to reconstruct an image from the cropped person regions, it often encounters the copy-paste issue during inference, leading to lower quality.
Second, the model also suffers from inferior text controllability.
The generated result simply combines multiple persons into one image, regardless of whether it is aligned with the input text.
In this work, we propose \approach{} to tackle this challenging task in a training-free manner.
Since the existing single-ID customization models have less copy-paste issue, our key idea is to adapt these models to achieve multi-ID customization.
To this end, we present an ID-decoupled cross-attention mechanism, injecting distinct ID embeddings into the corresponding image regions and thus generating multi-ID outputs.
To enhance the generation controllability, we introduce three critical strategies, namely the local prompt, depth-guided spatial control, and extended self-attention, making the results more consistent with the text prompts and ID images.
We also carefully build a benchmark, called \benchmark{}, for evaluation.
The extensive qualitative and quantitative results demonstrate the effectiveness of \approach{} in solving the aforementioned two challenges.
Its performance is comparable or even better than the training-based multi-ID customization methods.

\end{abstract}    
\section{Introduction}
\label{sec:intro}

\begin{figure}[t]
    \centering
    \includegraphics[width=\linewidth]{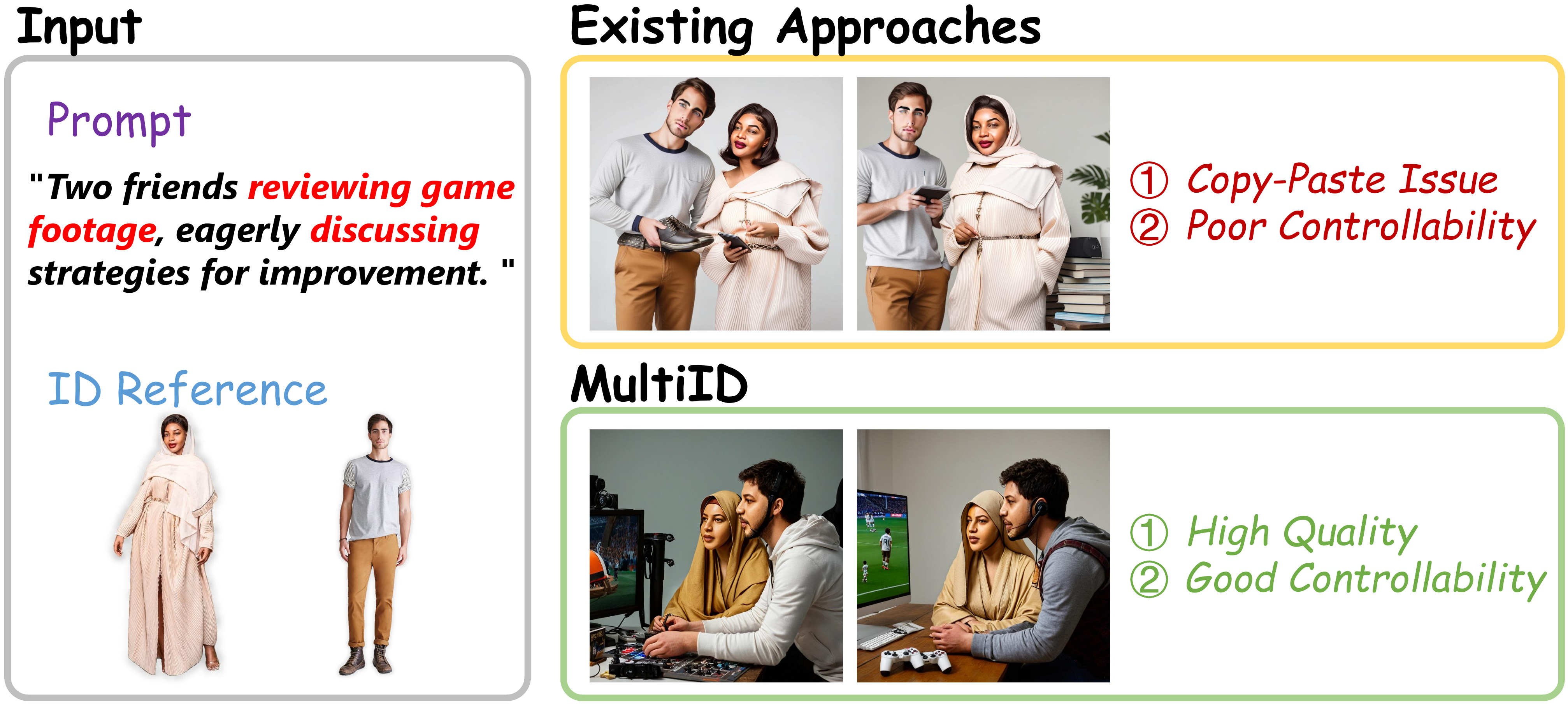}
    \caption{\approach{} shows better performance than existing methods in terms of generation quality and controllability.}
    \label{fig:teaser}
\end{figure}

Large text-to-image diffusion models, such as DALL-E~\cite{betker2023improving}, Stable Diffusion~\cite{rombach2022high, esser2024scaling}, and Flux~\cite{flux2023}, have brought significant advancements to image content creation.
Their superior generation quality and high controllability open up new prospects for a wide range of applications~\cite{brooks2023instructpix2pix, kawar2023imagic, ruiz2023dreambooth, poole2022dreamfusion, chen2023text2tex}.
Among these applications, person photo customization~\cite{li2024photomaker, guo2024pulid, zhou2024storymaker, wang2024instantid} has emerged as a compelling topic.
Given images of one or multiple persons as identity (ID) reference, its goal is to generate novel ID-preserving images that align with specific scenes, poses, and actions described in text prompts (see Figure~\ref{fig:teaser}). 
Additionally, the generated personalized images should maintain a quality comparable to that of vanilla image generation.

Currently, single-ID customization has been widely explored and achieved excellent results~\cite{li2024photomaker, cui2024idadapter}.
However, multi-ID customization, which aims to combine multiple person images into a coherent customized image, is less studied and poses unique challenges.
The first and biggest challenge is the serious copy-paste issue.
Since multi-ID customization does not have natural available datasets as single-ID customization, the previous methods~\cite{zhou2024storymaker, he2024uniportrait} typically build the dataset by cropping individual ID references from target multi-person images.
Hence, the model will converge to generate results through simply copying and pasting (see Figure~\ref{fig:teaser}), which harms the diversity and quality of personalized images.
Second, existing multi-ID customization methods have significant degradation in text controllability.
For example, as shown in Figure~\ref{fig:teaser}, the personalized images of existing approaches do not include the specified content such as ``reviewing game footage", ``discussing", and so on.

In this work, we introduce \approach{}, a novel \textbf{training-free} approach to tackling the challenging \textbf{multi-ID customization} task.
To address the \textbf{copy-paste issue} introduced by suboptimal datasets, our key idea is to leverage a pre-trained single-ID customization model as the backbone, and adapt it to the multi-ID scenario. 
Since the pre-trained model is trained on the high-quality dataset~\cite{li2024photomaker}, it inherently mitigates copy-paste issues and maintains high output quality.
To achieve this, we draw inspiration from a recent regional prompting method~\cite{omost} and present an \textit{ID-decoupled cross-attention} mechanism.
Specifically, we employ an ID encoder to encode multiple ID references separately and obtain their respective ID embeddings.
Next, we adjust the cross-attention layer of the pre-trained single-ID customization model so that each ID embedding only attends to specific pixels in the generated image and affects these pixels, while remaining invisible to others. 
By decoupling multiple IDs and acting on different regions of the image, \approach{} effectively extends the single-ID model to perform multi-ID customization.
It is worth noting that the entire process is training-free and does not require any parameter updates.

To enhance the controllability, we propose three effective strategies in \approach{}. 
First, in addition to the regular global prompt, we assign a local prompt for each ID to depict detailed information, such as clothing, action, posture, etc. 
The local prompts will be injected into their corresponding image regions using a mechanism similar to ID-decoupled cross-attention, thereby achieving precise and fine-grained control cover each ID.
Second, we leverage the powerful text-to-image diffusion model~\cite{flux2023} and ControlNet~\cite{zhang2023adding} to improve image-text consistency by incorporating depth-guided spatial control into \approach{}.
Specifically, we first employ Flux~\cite{flux2023} to generate a high-quality initial image that aligns well with the given global and local prompts.
This image is then transformed into a depth map, which serves as input to ControlNet~\cite{zhang2023adding}, providing additional spatial guidance for multi-ID customization.
As the depth map inherits some semantic information from the initial image, it is very helpful in reinforcing alignment with the text prompt.
The third strategy involves using extended self-attention to enhance ID consistency.
By inverting ID images through DDIM inversion~\cite{song2020denoising} and caching the intermediate features, we enable the self-attention layers to incorporate ID-specific information, thereby improving the ID consistency between the generated image and the reference ID images.


To verify the effectiveness of~\approach{}, we built a benchmark based on this challenging task, called~\benchmark{}.
\benchmark{} contains about 400 test samples, each of which includes images and local prompts of multiple IDs, as well as a global prompt.
Specifically, the ID images are mainly sampled from an open-source high-quality full-body human image dataset~\cite{fu2022stylegan}.
We generate the local and global prompts via LLMs, and manually check the benchmark to ensure diversity and accuracy.
To evaluate the generation quality, \benchmark{} measures the customized images from 3 perspectives: local prompt alignment, global prompt alignment, and ID alignment.
The extensive qualitative and quantitative results on~\benchmark{} demonstrate the effectiveness of~\approach{} in multi-ID personalization.
It successfully addresses the copy-paste issue and enhances controllability when compared with the baseline models.
In addition, we demonstrate that~\approach{} enables a novel application called background-preserving ID customization, which maintains background consistency with a reference image, further enhancing the practicality of~\approach{}.

\section{Related Work}
\label{sec:related_work}

\subsection{Controllable Text-to-Image Generation}
Previous methods, especially Generative Adversarial Networks (GANs) \cite{reed2016generative,zhang2017stackgan,brock2018large,li2019controllable,Xu_2021_CVPR}, provide a reliable outlook in the field of image generation.
Recent Diffusion models \cite{ho2020denoising,song2020denoising,nichol2021improved}, represented by Latent Diffusion \cite{rombach2022high}, Stable Diffusion XL \cite{podell2023sdxl}, Diffusion Transformer \cite{peebles2023scalable}, etc., showed state-of-the-art text-to-image generation performance, enhancing the significance of conditional image generation.
On these bases, multitudinous methods focusing on generation controllability emerged.
ControlNet \cite{zhang2023adding} introduces a flexible architecture for injecting various kinds of 2D conditions into pre-trained models.
MasaCtrl \cite{cao2023masactrl} adjusts the key and value in attention machines, achieving consistent image synthesis and editing.
Our method pays close attention to reliability and conceptual controllability under a comprehensive view of image generation pipelines.

\subsection{Subject-Driven Image Generation}
It's crucial to control a specific concept in generated images among various practical applications.
There are tuning-based subject-driven generation approaches that aim to capture detailed subject information from a small training set of related images \cite{gal2022image,ruiz2023dreambooth,kumari2023multi,gu2024mix}, or adapt to particular concepts through optimization \cite{sun2024rectifid,yao2024concept,wang2024spotactor,chen2025freecompose}. 
Recent tuning-free methods tend to reduce test-time costs by learning plug-in modules \cite{yan2023facestudio,ma2024subject,wang2024instantid,cui2024idadapter}. 
For typical examples, FasterComposer \cite{xiao2024fastcomposer}, IPAdapter \cite{ye2023ip}, PhotoMkaer \cite{li2024photomaker}, etc., encode user-provided images into embeddings that can be absorbed by pre-trained diffusion models through learned encoders, achieving efficient while consistent generation with very little extra inference consumption.
Some other approaches propose attention manipulations thus making results meet the personalized requirements \cite{cao2023masactrl,wang2024magicface,wang2024taming,chen2024training,liu2025training}.    
Our approach lies in a tuning-free setting, obtaining a general design that is data-independent and transferable.

\subsection{Multi-ID Personalization}
When extending the subject-driven image generation to customize multiple humans, tasks become complicated with the involvement of ID correspondence \cite{xiao2024fastcomposer}.
Tuning-based methods always confront ID blending problems since multi-concept joint training is acceptable to high-level semantic distinguishment but raises confusion for different subjects with the same concept category \cite{kumari2023multi}.
A typical solution is encoding specific subjects to tokens and directly integrating identity information into text embeddings \cite{avrahami2023break,jang2024identity,he2024uniportrait,li2024photomaker,zhou2024storymaker}. 
Still, it suffers from inconsistency between the compact embedding representation and pixel-wise 2D features.
Other approaches injecting subjects into generated images through pre-defined layout guidance \cite{zheng2023layoutdiffusion,sarukkai2024collage,wang2024ms,gu2024mix,kim2024instantfamily,kwon2024concept,wang2024spotactor,kong2025omg} often lack customization or interaction of multiple concepts.
Recent methods adopt manually designed attention operations for local controllability and rational interactions \cite{zhou2024migc,hoe2024interactdiffusion,wang2024moa}. 
Nevertheless, bringing myriad parallel control objectives into attention mechanisms inevitably makes it difficult to balance image quality and controllability, and makes it easy to fall into a copy-paste solution.
To this end, we conduct interaction generation and identity control sequentially, solving complex problems with a concise and clear framework.
Thus, our proposed framework supports high-quality multi-ID personalization while addressing ID blending and interaction deficiency.

\section{Methodology}

\begin{figure*}
    \centering
    \includegraphics[width=\linewidth]{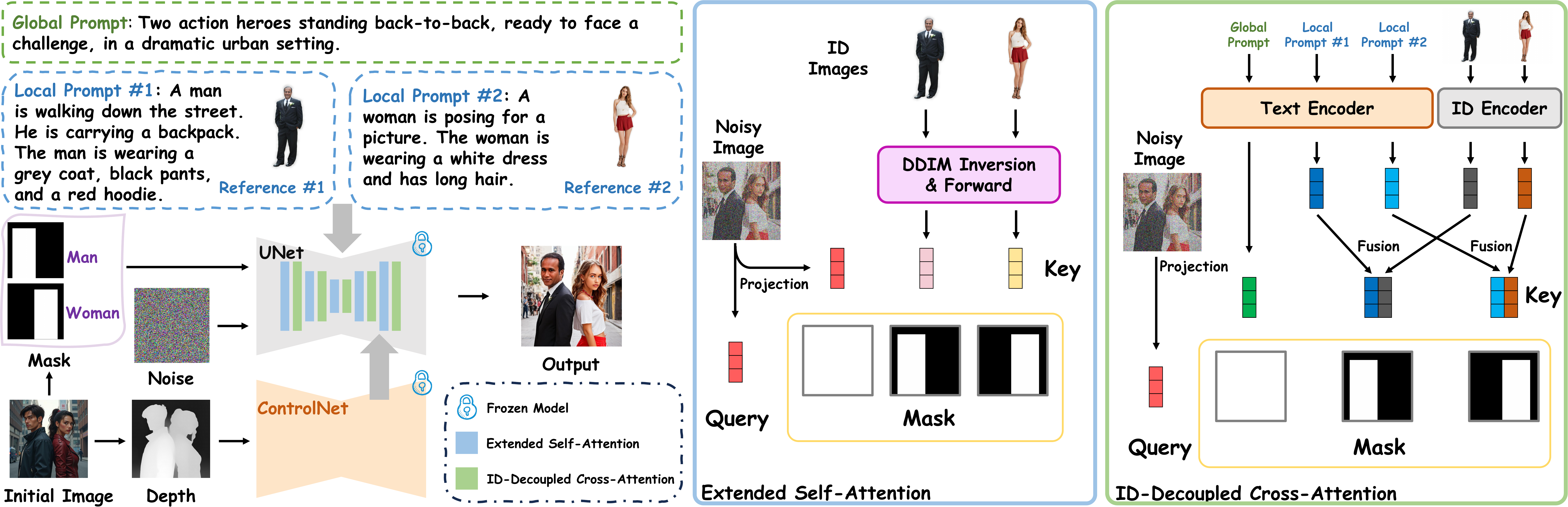}
    \caption{Illustration of our proposed \approach{}. \approach{} takes a global prompt, multiple local prompts and ID images as input, and produces a personalized image accordingly. It consists of three critical components, termed (1) depth-guided spatial control, (2) extended self-attention, and (3) ID-decoupled cross-attention.}
    \label{fig:method}
\end{figure*}

In this section, we elaborate on \approach{} (see Figure~\ref{fig:method}), a \textbf{training-free} method for multi-ID customization.
In what follows, we will first introduce the text-to-image (T2I) diffusion model, especially its attention mechanism on which our method is based (Section~\ref{sec:pre}).
We then propose transforming the existing single-ID personalization model into a multi-ID model.
This is achieved by decoupling multiple ID and text features and injecting them into different regions of the generated image through the cross-attention operation (Section~\ref{sec:CA}).
In addition, we further enhance the generation quality and consistency by introducing depth-guided spatial control (Section~\ref{sec:depth_control}) and extended self-attention (Section~\ref{sec:SA}) strategies.

\subsection{Preliminary}
\label{sec:pre}

T2I diffusion models have emerged as a powerful class of generative models in image synthesis.
These models start by drawing samples from a standard Gaussian distribution, gradually predicting and removing noise, ultimately producing high-quality image outputs that conform to input text prompts.
Typically, this is achieved by a denoising UNet~\cite{ronneberger2015u} with multiple blocks of self-attention and cross-attention layers.
More specifically, the self-attention layer operates on image features, while the cross-attention layer operates between image and text features.
Given the image features, denoted as $X$, the self-attention layer first projects them into queries ($Q^s$), keys ($K^s$) and values ($V^s$) via three linear transformations:
\begin{equation}
Q^s = X \cdot W_Q^s, \ K^s = X \cdot W_K^s, \ V^s = X \cdot W_V^s.
\end{equation}
Here, $W_Q^s, W_K^s,$ and $W_V^s$ are learnable projection matrices. 
Then, the self-attention map $A^s$ between the queries and keys can be calculated as:
\begin{equation}
\label{eq:attention}
    A^s = \text{Softmax} (\frac{Q^s{K^s}^T}{\sqrt{d}}),
\end{equation}
where $d$ represents the dimension of key vectors. 
The self-attention output $H^s$ is a weighted sum of the value vectors using the weights in the self-attention map, i.e.,
\begin{equation}
\label{eq:output}
H^s = A^sV^s.
\end{equation}

The cross-attention layer injects text conditions into image synthesis process through a similar attention mechanism between image features $X$ and text features $P$.
In the implementation, the text features are typically obtained from a pre-trained CLIP~\cite{radford2021learning} text encoder, which can effectively encode input prompts into semantic representations.
Another three projection matrices, $W_Q^c, W_K^c,$ and $W_V^c$, are used here to compute the queries ($Q^c$), keys ($K^c$), and values ($V^c$) for the cross-attention operation: $Q^c = X \cdot W_Q^c, \ K^c = P \cdot W_K^c, \ V^c = P \cdot W_V^c$.
Then, the cross-attention map $A^c$ and output $H^c$ are computed following a similar procedure as in Equations~\ref{eq:attention} and~\ref{eq:output}.

\subsection{Overview}
\label{sec:overview}

We develop \approach{} for multi-ID personalization.
Given $N$ different ID images $\{e_i\}_{i=1}^N$ as reference, where $e_i$ represents the $i$-th identity, \approach{} combines them together to generate an identity-preserving and harmonized image output $x$ (see Figure~\ref{fig:method}).
Additionally, the text prompts can be specified to further enhance the generation controllability.
In this work, we consider two types of text prompts.
One is called the global prompt $p^g$, which describes the overall semantics of the generated image (e.g., background, style).
The other is the local prompts $\{p^l_i\}_{i=1}^N$, which depict fine-grained attributes associated with individuals (e.g., expression, pose). 
Here, $p^l_i$ denotes the local prompt for the $i$-th ID. 
More formally, \approach{} can be formulated as a mapping $f$, such that $x = f(\{e_i\}_{i=1}^N,  p^g,  \{p^l_i\}_{i=1}^N)$.
It is worth noting that the $f$ used in this work is originally designed for single-ID customization.
We extend it to the multi-ID setting in a \textit{training-free} manner, enabling identity fusion without modifying any model parameters.

\subsection{ID-Decoupled Cross-Attention}
\label{sec:CA}

As illustrated in Figure~\ref{fig:teaser}, existing multi-ID customization methods often suffer from severe copy-paste issues and limited controllability.
Hence, in this work, we propose to build \approach{} upon a pre-trained single-ID model (e.g., PhotoMaker-v2~\cite{li2024photomaker}).
The first step here is to extend the model's capability to support multi-ID input.
To achieve this, we draw inspiration from a recent regional prompting method, called Omost~\cite{omost}.
By manipulating the scope of cross-attention operation through masking, it enables the injection of multiple text prompts into distinct image regions, ensuring a harmonized generation result that integrates various local conditions.
Following the idea, we present an ID-decoupled cross attention strategy, seamlessly applying the single-ID model to the multi-ID setting in a training-free manner.
To be more specific, we first convert the global prompt $p^g$ and each identity's local prompt $p^l_i$ into text embeddings using a pre-trained CLIP~\cite{radford2021learning} text encoder $\phi$:
\begin{equation}
    P^g = \phi(p^g), \ \tilde{P^l_i} = \phi(p^l_i).
\end{equation}
The ID images are also encoded into visual embeddings and fused with their corresponding text embeddings to construct a distinctive feature representation for each individual:
\begin{equation}
    E_i = \psi(e_i), \ P^l_i = \texttt{fusion}(\tilde{P^l_i}, E_i).
\end{equation}
Here, $\psi$ is an ID encoder, \texttt{fusion} refers to the fusion operation between an ID embedding and a text embedding, and $P^l_i$ denotes the resulting feature for the $i$-th identity.
The proposed ID-decoupled cross-attention mechanism restricts the decoupled ID features to specific spatial regions via a masked cross-attention procedure, i.e.,
\begin{align}
    P &= [P^g \oplus P^l_1 \oplus \dots \oplus P^l_N], \notag \\
    K^c &= P \cdot W_K^c, \ V^c = P \cdot W_V^c, \notag \\
    M^c &= [\mathbf{1} \oplus M_1 \oplus \dots \oplus M_N], \\
    A^c &= \text{Softmax} (Q^c{K^c}^T / \sqrt{d} + \log M^c), \notag \\
    H^c &= A^cV^c, \notag
\end{align}
where $M_i$ indicates the $i$-th ID's attention mask (obtained from the bounding box introduced in Section~\ref{sec:idbench}, with a value of 1 inside the box and 0 outside the box), and $\oplus$ denotes the concatenation operation.
Notably, the global prompt corresponds to an all-1 attention mask, which means that all image pixels can attend to this prompt and are influenced by it.
In contrast, the local prompts and ID images can only interact and guide image generation in certain regions, thus elegantly achieving multi-ID customization with a pre-trained single-ID model.

\subsection{Depth-Guided Spatial Control}
\label{sec:depth_control}

Although the introduction of ID-decoupled cross-attention mechanism makes multi-ID customization tractable, it still presents some limitations.
First, the initial implementation sometimes causes aliasing issues, where different characters share body parts, particularly when their bounding box regions are closely positioned.
Second, its generated results may not always align with the posture, action, and so on that are specified in the text prompt, leading to inconsistency.

To overcome these limitations, our intuition is as follows.
For the aliasing issue, it seems that we need to introduce stronger spatial constraints in the generation process.
In terms of inconsistency, our idea is using a more accurate initial image to guide the generation.
To this end, we develop a simple yet effective strategy, termed depth-guided spatial control, within \approach{}.
Specifically, we first leverage the powerful image generator Flux~\cite{flux2023} to produce an initial image based on both global and local prompts.
We then extract a depth map from this initial image and feed it into ControlNet~\cite{zhang2023adding}, which enforces spatial constraints, ensuring precise positioning of individuals in the multi-ID customization process without any aliasing (see Figure~\ref{fig:method}).
Furthermore, thanks to Flux’s outstanding text-image consistency, the depth map inherits some semantic information from the initial image (e.g., posture, action), which also enhances the generation consistency between the customized image and the text prompts.

\subsection{Extended Self-Attention}
\label{sec:SA}

Another critical technique of \approach{} is called extended self-attention, which is present to further improve the consistency with the ID reference images.
The core idea is to allow the customized image to additionally attend to the ID images in the self-attention layers.
Since the ID images are all realistic, we perform DDIM~\cite{song2020denoising} inversion to map each ID image $e_i$ to a latent vector and cache its image features $X_i$ (omitting the self-attention layer index for brevity) during the forward process from the inverted latent.
The extended self-attention is then formulated as follows:
\begin{align}
    K^s &= [X \oplus X_1 \oplus \dots \oplus X_N] \cdot W_K^s, \notag \\
    V^s &= [X \oplus X_1 \oplus \dots \oplus X_N] \cdot W_V^s, \notag \\
    M^s &= [\mathbf{1} \oplus M_1 \oplus \dots \oplus M_N], \\
    A^s &= \text{Softmax} (Q^s{K^s}^T / \sqrt{d} + \log M^s), \notag \\
    H^s &= A^sV^s. \notag
\end{align}
Similarly, we use the attention mask to restrict the extended self-attention to occur only between corresponding IDs.

\section{Experiments}

\subsection{Setup}

\noindent\textbf{Baselines.}
For multi-ID personalization, we compare our method with previous state-of-the-art methods: FastComposer \cite{xiao2024fastcomposer}, UniPortrait \cite{he2024uniportrait}, and StoryMaker \cite{zhou2024storymaker}.
These methods are applicable to multiple reference identities, with the first two mostly focusing on facial consistency. 
In addition, we also compare InstanceDiffusion \cite{wang2024instancediffusion} which introduces text-based personalization by injecting personalized descriptions into layout conditions.

\noindent\textbf{Benchmark.}
We evaluate the performance of multi-ID personalization on our proposed \benchmark{}.
\benchmark{} consists of 40 kinds of human interactions and each kind contains 10 prompts.
Since we excluded 7 low-quality samples, we have 393 different samples in total. 
We generate 4 images for each sample and report the average metrics of 1572 generated images.
More details about \benchmark{} are provided in the supplementary material.

\noindent\textbf{Implementation.}
We evaluate baselines and our proposed method on \benchmark{}, generating 4 images for each sample and calculating the average metrics as the final result.
For FastComposer and UniPortrait, which focus on facial consistency, we concatenate the ID descriptions behind prompts to ensure fair and sufficient input information.
For InstanceDiffusion, we directly utilize the bounding boxes detected during benchmark construction as the layout guidance, while injecting the corresponding ID descriptions as conditions for personalization.


\subsection{Evaluation Metrics}

\noindent\textbf{Global Generation Quality.}
We utilize CLIP-B/32 \cite{radford2021learning} text-to-image similarity to compare generation consistency (CLIP-T).
We also adopt the Human Preference Score v2 (HPSv2) \cite{wu2023human} as a quality standard for generation.


\noindent\textbf{ID Alignment.}
To validate the local alignment of IDs, we first detect humans in generated images using Grounding DINO \cite{liu2023grounding}.
Then, we utilize a greedy algorithm to match cropped person images with reference images through their CLIP image feature similarities.
ID alignment consists of image-level whole-body consistency and face similarity.
We adopt CLIP-I \cite{gal2022image} for matched image pairs to evaluate whole-body consistency (Body).
Subsequently, we detect faces from these person images using RetinaFace \cite{Deng2020CVPR} and extract their embeddings using FaceNet \cite{schroff2015facenet}, thus applying these embeddings to calculate face similarity (Face).

\noindent\textbf{Local Prompt Alignment.}
Profited from that \benchmark{} includes descriptions of both posture and appearance, we validate local text-to-image alignment from two perspectives: overall alignment and posture accuracy, thus evaluating local generation quality and accuracy from aspects of person appearance and motion.
We adopt CLIP-T for these two metrics, utilizing local appearance + posture descriptions for full local prompt alignment (Full) and solely posture descriptions for posture consistency (Pose).  
This design helps to check for the copy-paste issue as it inflates general ID similarity metrics but is helpless about posture consistency.
Meanwhile, it can reflect the method's capability to handle complex samples with complicated human status.

\subsection{Results}

\begin{figure*}
    \centering
    \includegraphics[width=0.99\linewidth]{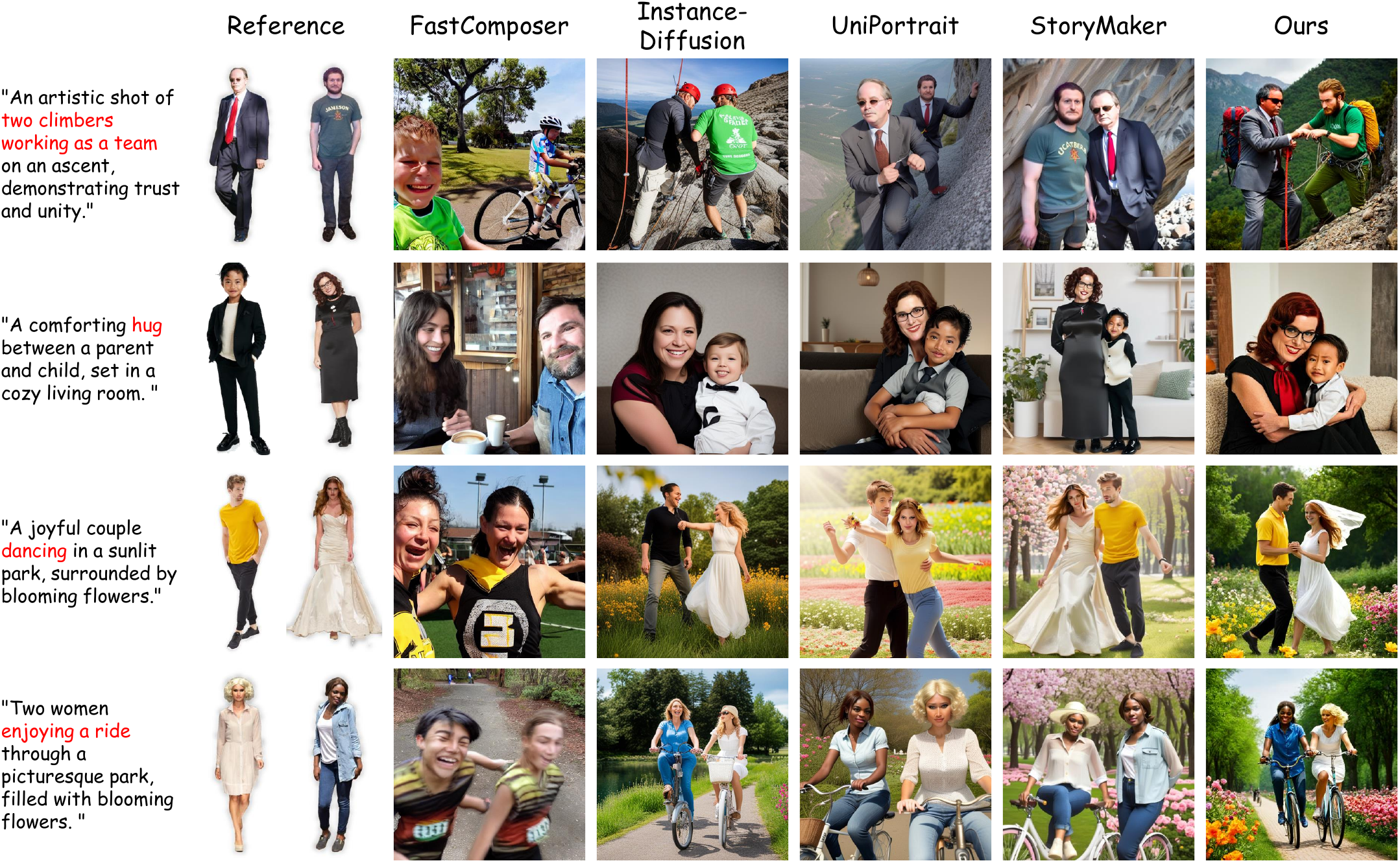}
    \caption{
    Qualitative comparison of different methods on \benchmark{}.
    We {\color{red} highlight} the description of human interactions in prompts, emphasizing the comparison of human postures in the generated images.
    }
    \label{fig:comparison}
\end{figure*}

\begin{table*}[!ht]
    \centering
    \begin{tabular}{l|cc|cc|cc}
    
    
    \toprule
        \multirow{2}{*}{Method} & \multicolumn{2}{c|}{\textit{Global Align.}} & \multicolumn{2}{c|}{\textit{ID Align.}} & \multicolumn{2}{c}{\textit{Local Prompt Align.}} \\
        ~ & CLIP-T $(\%)$ & HPSv2 $(\%)$ & Body $(\%)$ & Face $(\%)$ & Full $(\%)$ & Pose $(\%)$ \\ 
    \midrule
        FastComposer \cite{xiao2024fastcomposer} & \colorize{21.9}{28.4}{21.9} & \colorize{21.0}{29.3}{21.0} & \colorize{47.9}{72.1}{47.9} & \colorize{1.8}{33.9}{1.8} & \colorize{18.1}{29.6}{18.1} & \colorize{18.7}{26.6}{18.7} \\ 
        InstanceDiffusion \cite{wang2024instancediffusion} & \secondmark{\colorize{21.9}{28.4}{28.0}} & \secondmark{\colorize{21.0}{29.3}{27.3}} & \colorize{47.9}{72.1}{62.8} & \colorize{1.8}{33.9}{2.3} & \secondmark{\colorize{18.1}{29.6}{27.5}} & \secondmark{\colorize{18.7}{26.6}{26.1}} \\  
        UniPortrait \cite{he2024uniportrait} & \colorize{21.9}{28.4}{26.5} & \colorize{21.0}{29.3}{27.2} & \colorize{47.9}{72.1}{64.0} & \colorize{1.8}{33.9}{20.4} & \colorize{18.1}{29.6}{27.0} & \colorize{18.7}{26.6}{25.3} \\ 
        StoryMaker \cite{zhou2024storymaker} & \colorize{21.9}{28.4}{26.0} & \colorize{21.0}{29.3}{26.8} & \bestmark{\colorize{47.9}{72.1}{72.1}} & \bestmark{\colorize{1.8}{33.9}{33.9}} & \colorize{18.1}{29.6}{25.5} & \colorize{18.7}{26.6}{22.9} \\
    \hline
        \textbf{Ours} & \bestmark{\colorize{21.9}{28.4}{28.4}} & \bestmark{\colorize{21.0}{29.3}{29.3}} & \secondmark{\colorize{47.9}{72.1}{64.1}} & \secondmark{\colorize{1.8}{33.9}{21.2}} & \bestmark{\colorize{18.1}{29.6}{29.6}} & \bestmark{\colorize{18.7}{26.6}{26.6}} \\ 
    \bottomrule
    
    \end{tabular}
    \caption{
    Quantitative comparison of different methods on \benchmark.
    For \textit{Global Alignment}, CLIP-T and HPSv2 indicate text-to-image similarities between the global prompt and generated image. 
    For \textit{ID Alignment}, Body and Face respectively represent the similarity between the CLIP features of the body / the FaceNet embeddings of the face and the corresponding reference image.
    For \textit{Local Prompt Alignment}, Full and Pose refer to the CLIP-T between the local description / the local posture description and the ID crop.
    The higher these metrics are, the better the corresponding method performs.
    The best and second best results are \bestmark{bolded} and \secondmark{underlined}, respectively. 
    Cells are highlighted from \colorbox{mygreen}{lower} to \colorbox{myred}{higher}.
    }
\label{tab:comparison}
\end{table*}

\newcommand{\diffg}[1]{{\color{red}$_{(+#1)}$}}
\newcommand{\diffl}[1]{{\color{blue}$_{(-#1)}$}}

\begin{table*}[!ht]
    \centering
    \begin{tabular}{ccc|cc|cc|cc}
    
    
    \toprule
        \multicolumn{3}{c|}{Method} & \multicolumn{2}{c|}{\textit{Global Align.}} & \multicolumn{2}{c|}{\textit{ID Align.}} & \multicolumn{2}{c}{\textit{Local Prompt Align.}} \\
        D.C. & S.A. & C.A. & CLIP-T $(\%)$ & HPSv2 $(\%)$ & Body $(\%)$ & Face $(\%)$ & Full $(\%)$ & Pose $(\%)$ \\  
    \midrule
        ~ & ~ & ~ & \colorize{26.9}{28.9}{27.1} \diffl{1.3} & \colorize{26.5}{30.2}{26.5} \diffl{2.8} & \colorize{46.9}{64.1}{46.9} \diffl{17.2} & \colorize{11.5}{21.2}{11.5} \diffl{9.7} & \colorize{21.2}{29.6}{21.2} \diffl{8.4} & \colorize{19.8}{26.6}{19.8} \diffl{6.8} \\ 
        \checkmark & ~ & ~ & \bestmark{\colorize{26.9}{28.9}{28.9}} \diffg{0.5} & \secondmark{\colorize{26.5}{30.2}{30.2}} \diffg{0.9} & \colorize{46.9}{64.1}{59.7} \diffl{4.4} & \colorize{11.5}{21.2}{13.9} \diffl{7.3} & \colorize{21.2}{29.6}{27.4} \diffl{2.2} & \colorize{19.8}{26.6}{25.9} \diffl{0.7} \\ 
        ~ & \checkmark & \checkmark & \colorize{26.9}{28.9}{26.9} \diffl{1.5} & \colorize{26.5}{30.2}{26.4} \diffl{2.9} & \colorize{46.9}{64.1}{55.1} \diffl{9.0} & \colorize{11.5}{21.2}{13.2} \diffl{8.0} & \colorize{21.2}{29.6}{24.0} \diffl{5.6} & \colorize{19.8}{26.6}{21.5} \diffl{5.1} \\ 
        \checkmark & ~ & \checkmark & \colorize{26.9}{28.9}{28.5} \diffg{0.1} & \colorize{26.5}{30.2}{29.1} \diffl{0.1} & \secondmark{\colorize{46.9}{64.1}{62.3}} \diffl{1.8} & \secondmark{\colorize{11.5}{21.2}{20.0}} \diffl{1.2} & \secondmark{\colorize{21.2}{29.6}{28.9}} \diffl{0.7} & \colorize{19.8}{26.6}{26.0} \diffl{0.6} \\ 
        \checkmark & \checkmark & ~ & \secondmark{\colorize{26.9}{28.9}{28.9}} \diffg{0.5} & \bestmark{\colorize{26.5}{30.2}{30.2}} \diffg{0.9} & \colorize{46.9}{64.1}{61.3} \diffl{2.8} & \colorize{11.5}{21.2}{15.1} \diffl{6.1} & \colorize{21.2}{29.6}{27.9} \diffl{1.7} & \secondmark{\colorize{19.8}{26.6}{26.4}} \diffl{0.2} \\ 
    \hline
        \checkmark & \checkmark & \checkmark & \colorize{26.9}{28.9}{28.4} & \colorize{26.5}{30.2}{29.3} & \bestmark{\colorize{46.9}{64.1}{64.1}} & \bestmark{\colorize{11.5}{21.2}{21.2}} & \bestmark{\colorize{21.2}{29.6}{29.6}} & \bestmark{\colorize{19.8}{26.6}{26.6}} \\ 
    \bottomrule
    
    \end{tabular}
    
    \caption{
    Ablation studies for main components of \approach.
    D.C., S.A., and C.A. indicate Depth Control, Extended Self-Attention, and ID-Decoupled Cross-Attention, respectively.
    We mark the metric margins in ablation experiments (greater with {\color{red}red} and less with {\color{blue}blue}).
    The best and second best results are \bestmark{bolded} and \secondmark{underlined}, respectively.
    Cells are highlighted from \colorbox{mygreen}{lower} to \colorbox{myred}{higher}.
    }
\label{tab:ablation}
\end{table*}

\begin{figure*}
    \centering
    \includegraphics[width=0.95\linewidth]{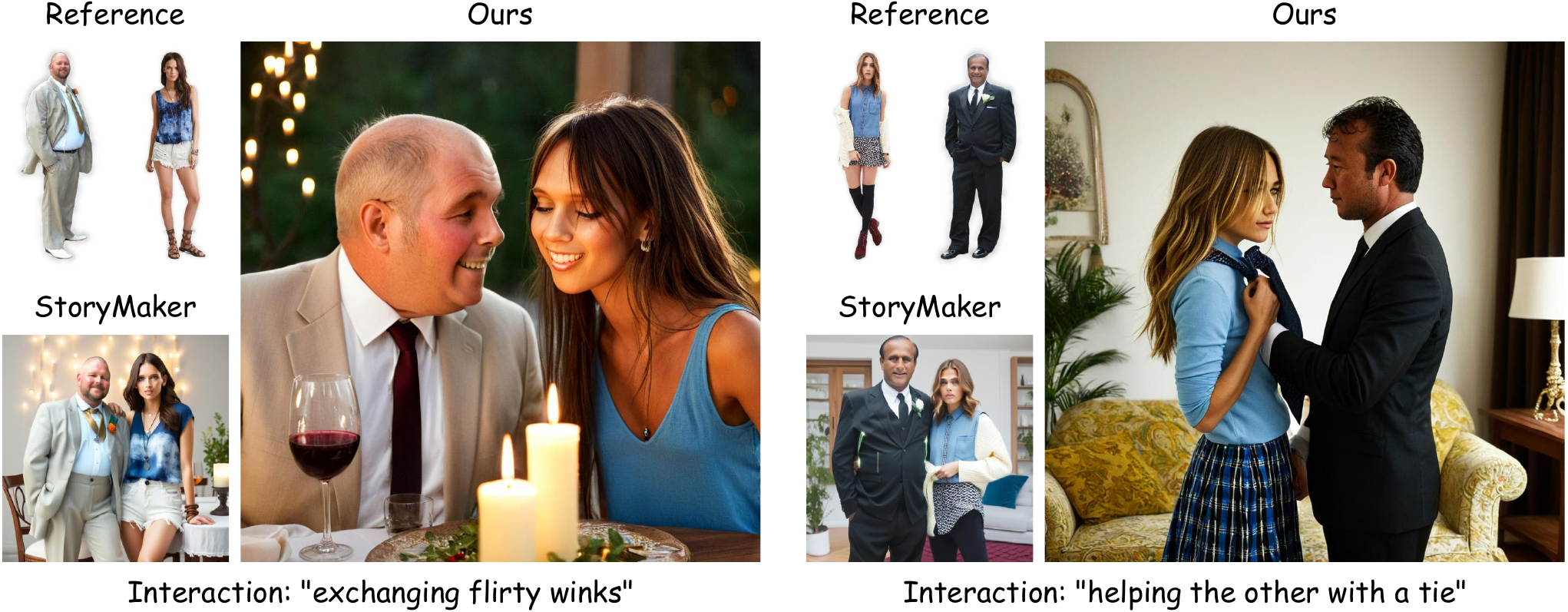}
    \caption{
    Visualization results of complex interactions.
    }
    \label{fig:interaction}
\end{figure*}

As shown in Table \ref{tab:comparison}, our proposed approach manifests a praiseworthy balance between prompt consistency and identity preservation.
\approach{} records the highest score in global text-to-image alignment (CLIP-T), human preference (HPSv2), ID text-to-image alignment (Full), and posture consistency (Pose), claims its capability to generate high-quality images that conform to requirements from general to local details.
Simultaneously, our method achieves competitive performance in ID image-to-image alignment (Body \& Face), demonstrating its reliability for high-flexible control.
Notably, StoryMaker showcases highlighted results in image-to-image ID alignment but struggles with text-to-image alignment, especially in terms of posture.
This is due to the notorious \textit{copy-paste issue}, which is induced as an easier solution for personalization but leads to a mediocre ability to generate complex samples.

To elaborate, we present a visual comparison of the qualitative outcomes derived from various methods in Figure \ref{fig:comparison}.
\approach{} vividly integrates multiple identities into high-quality outputs according to the specified text prompts, demonstrating strong capability in multi-ID customization.
Compared to \approach{}, baseline methods either exhibit inferior ID consistency (e.g., FastComposer, InstanceDiffusion), suffer from poor text controllability (e.g., the first case of StoryMaker), or encounter severe copy-paste issues (e.g., StoryMaker).
Additionally, the training-free nature of \approach{} further underscores its superiority in multi-ID customization.


We also illustrate the controllability of our method by specifically analyzing samples with complex human interactions in prompts as shown in Fig.\ref{fig:interaction}. 
Existing multi-ID personalization approaches such as StoryMaker tend to rigidly composite reference identities to meet ID consistency requirements, indeed ensuring high identity fidelity to reference IDs but inevitably sacrificing controllability and generalizability.
In contrast, our method employs ID-agnostic initial-image-based depth control and subsequent attention-based ID injection, which fully preserves the controllability of text-to-image diffusion models while maintaining the distinctiveness of individual subjects. 
It enables our framework to comprehensively address multi-ID personalized generation challenges, achieving a superior balance between identity preservation and compositional flexibility.


\subsection{Ablation Study}

\begin{figure}
    \centering
    \includegraphics[width=1\linewidth]{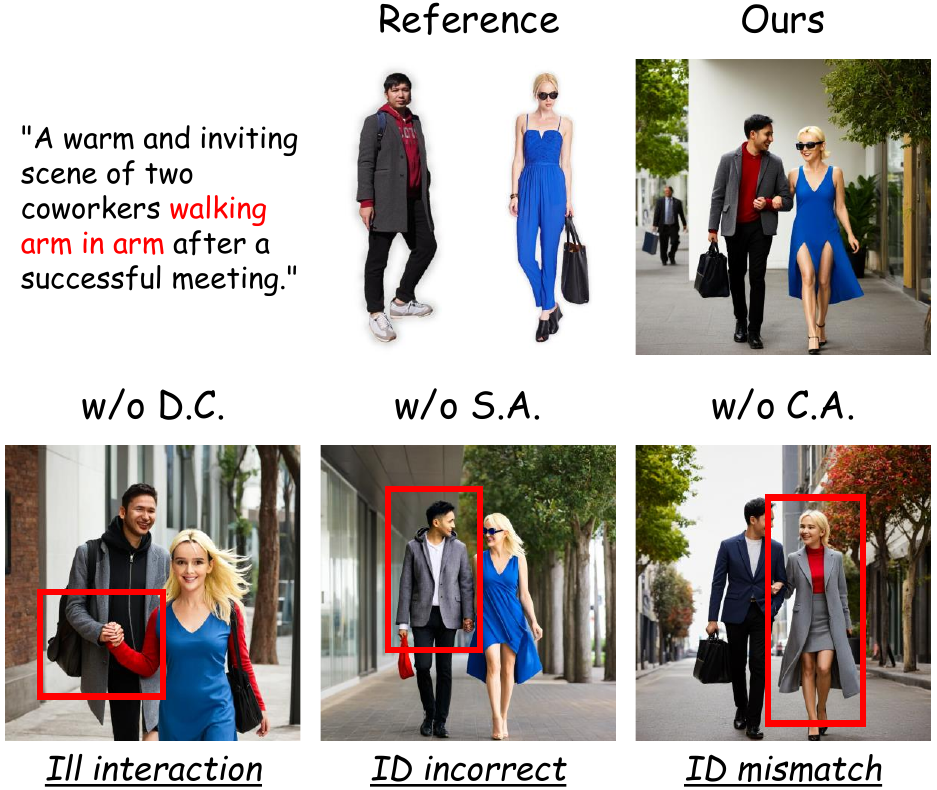}
    \caption{
    Qualitative analysis of ablation studies.
    We underline the issues below the images.
    \textit{\underline{Ill interaction}} indicates the interactions between IDs are unadjusted.
    \textit{\underline{ID incorrect}} denotes that the appearance of personalized ID is not inconsistent with its reference.
    \textit{\underline{ID mismatch}} represents the confusion arises between different IDs.
    The red boxes mark where the mentioned issues occur.
    }
    \label{fig:ablation}
\end{figure}

As shown in Tab.\ref{tab:ablation}, we evaluate the significance of each main component of our proposed \approach{} on \benchmark{}.
From a holistic metric perspective, our full framework exhibits marginal reductions in global text-image alignment compared to partial module configurations.
However, each component of our method plays a critical role in enhancing personalization capabilities. 
The initial-image-based depth control significantly improves generation fidelity and ID consistency. 
While introducing minor degradation in global alignment, the proposed attention manipulations substantially enhance ID preservation across different identities. 
Collectively, these components achieve an optimal trade-off between overall quality and personalization, demonstrating that strategic compromises in global alignment metrics can yield superiority in multi-ID personalization tasks.

Furthermore, the qualitative analysis in Fig.\ref{fig:ablation} substantiates the distinct roles of each component for personalization. 
Empirically, ill interaction (interactions between IDs are unadjusted), ID incorrect (inconsistent ID appearance), and ID mismatch (confusion between different IDs) are common issues in training-free multi-ID personalization tasks.
D.C. (Depth Control) enables natural and controllable human postures / interactions through identity-agnostic holistic guidance, where the absence of D.C. leads to physically implausible interactions (e.g., misaligned limb positions).
S.A. (Extended Self-Attention) directly injects reference ID appearance features, preventing significant deviations from reference images while maintaining prompt fidelity. 
C.A. (ID-Decoupled Cross-Attention) ensures regionally constrained ID injection via spatial masking, effectively eliminating the impact of identity mismatch (e.g., blended body features across IDs). 
This modular synergy demonstrates that decoupled control of contextual coherence, local appearance, and ID alignment is critical for robust multi-ID personalization.


\subsection{Applications}

\begin{figure}
    \centering
    \includegraphics[width=\linewidth]{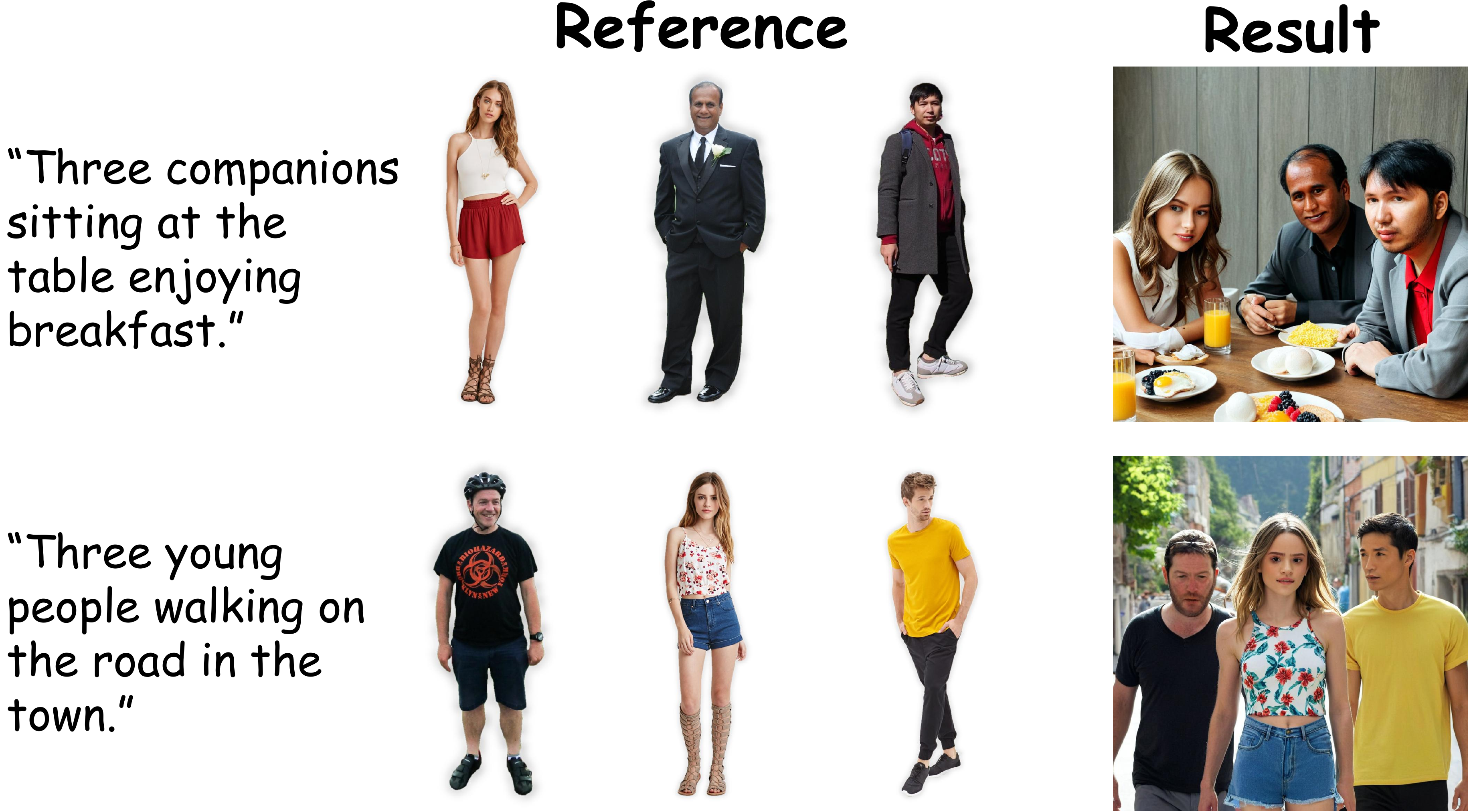}
    \caption{
    \approach{} can also perform ID customization for more than two persons.
    }
    \label{fig:multi_person}
\end{figure}

We apply \approach{} to ID customization for three persons, and show the generated images in Figure~\ref{fig:multi_person}.
From the results, we see that \approach{} successfully integrates three distinct IDs into a cohesive output, demonstrating that the proposed training-free method remains highly effective in such complex scenarios.

\begin{figure}
    \centering
    \includegraphics[width=\linewidth]{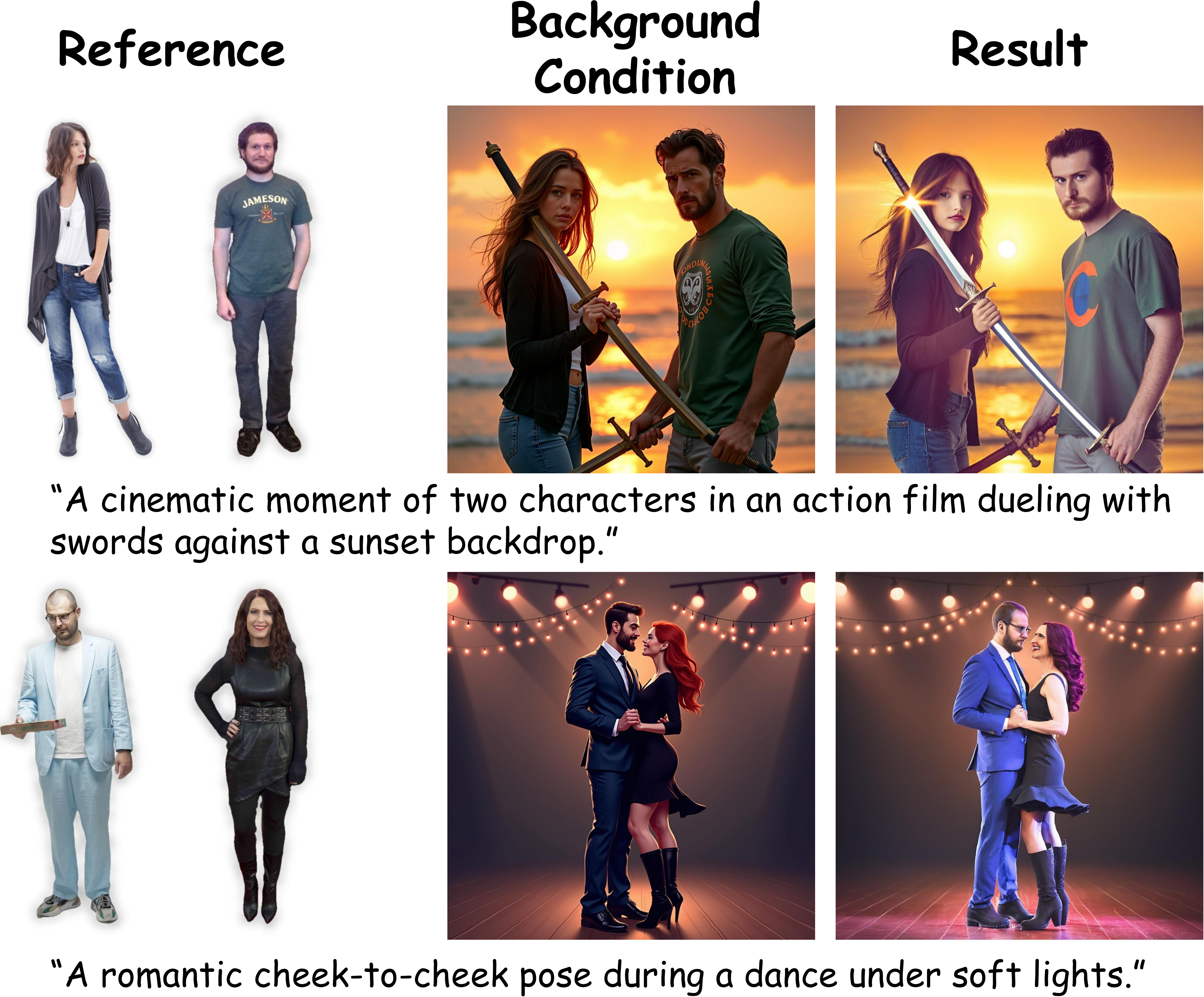}
    \caption{
    Application of \approach{} on background-preserving ID customization.
    }
    \label{fig:bg_preserve}
\end{figure}

Besides, we also present a novel application to preserve the background in customized images (see Figure~\ref{fig:bg_preserve}).
This is achieved through a strategy similar to RePaint~\cite{lugmayr2022repaintinpaintingusingdenoising}.
Specifically, we replace the predicted noise in the background region with the real noise obtained in the forward process, while keeping the predicted noise in the foreground region.
This innovative application enables users to upload a background reference and seamlessly replace the characters within it using specified IDs, achieving a compelling "cutout" effect.

\section{Conclusion}

This work studies multi-ID customization, which aims to generate ID-preserving customized images based on given multiple IDs.
To address the challenging task, we propose a training-free method, namely \approach{}.
By proposing an ID-decoupled cross-attention mechanism, we effectively adapt the pre-trained single-ID model to the multi-ID customization scenario and overcome the copy-paste issue.
Additionally, the introduction of local prompts, depth-guided spatial control, and extended self-attention enhances the generation controllability and consistency.
We also contribution a benchmark in this paper for pair comparison.
The quantitative and qualitative results demonstrate the superiority of \approach{} in terms of ID consistency and prompt alignment.
In the future, we will explore the potential of the proposed training-free framework in other difficult tasks, such as story generation across images.

\newpage
{
    \small
    \bibliographystyle{ieeenat_fullname}
    \bibliography{main}

@String(CVPR= {IEEE Conf. Comput. Vis. Pattern Recog.})

@String(ICCV= {Int. Conf. Comput. Vis.})

@String(CVPR  = {CVPR})

@String(ICCV  = {ICCV})

@article{betker2023improving,
  title={Improving image generation with better captions},
  author={Betker, James and Goh, Gabriel and Jing, Li and Brooks, Tim and Wang, Jianfeng and Li, Linjie and Ouyang, Long and Zhuang, Juntang and Lee, Joyce and Guo, Yufei and others},
  journal={Computer Science. https://cdn. openai. com/papers/dall-e-3. pdf},
  volume={2},
  number={3},
  pages={8},
  year={2023}
}

@inproceedings{esser2024scaling,
  title={Scaling rectified flow transformers for high-resolution image synthesis},
  author={Esser, Patrick and Kulal, Sumith and Blattmann, Andreas and Entezari, Rahim and M{\"u}ller, Jonas and Saini, Harry and Levi, Yam and Lorenz, Dominik and Sauer, Axel and Boesel, Frederic and others},
  booktitle={Forty-first International Conference on Machine Learning},
  year={2024}
}

@inproceedings{rombach2022high,
  title={High-resolution image synthesis with latent diffusion models},
  author={Rombach, Robin and Blattmann, Andreas and Lorenz, Dominik and Esser, Patrick and Ommer, Bj{\"o}rn},
  booktitle={Proceedings of the IEEE/CVF conference on computer vision and pattern recognition},
  pages={10684--10695},
  year={2022}
}

@misc{flux2023,
    author={Black Forest Labs},
    title={FLUX},
    year={2023},
    howpublished={\url{https://github.com/black-forest-labs/flux}},
}

@inproceedings{li2024photomaker,
  title={Photomaker: Customizing realistic human photos via stacked id embedding},
  author={Li, Zhen and Cao, Mingdeng and Wang, Xintao and Qi, Zhongang and Cheng, Ming-Ming and Shan, Ying},
  booktitle={Proceedings of the IEEE/CVF Conference on Computer Vision and Pattern Recognition},
  pages={8640--8650},
  year={2024}
}

@article{wang2024instantid,
  title={Instantid: Zero-shot identity-preserving generation in seconds},
  author={Wang, Qixun and Bai, Xu and Wang, Haofan and Qin, Zekui and Chen, Anthony and Li, Huaxia and Tang, Xu and Hu, Yao},
  journal={arXiv preprint arXiv:2401.07519},
  year={2024}
}

@article{guo2024pulid,
  title={Pulid: Pure and lightning id customization via contrastive alignment},
  author={Guo, Zinan and Wu, Yanze and Chen, Zhuowei and Chen, Lang and Zhang, Peng and He, Qian},
  journal={arXiv preprint arXiv:2404.16022},
  year={2024}
}

@article{zhou2024storymaker,
  title={Storymaker: Towards holistic consistent characters in text-to-image generation},
  author={Zhou, Zhengguang and Li, Jing and Li, Huaxia and Chen, Nemo and Tang, Xu},
  journal={arXiv preprint arXiv:2409.12576},
  year={2024}
}

@article{he2024uniportrait,
  title={UniPortrait: A Unified Framework for Identity-Preserving Single-and Multi-Human Image Personalization},
  author={He, Junjie and Geng, Yifeng and Bo, Liefeng},
  journal={arXiv preprint arXiv:2408.05939},
  year={2024}
}

@inproceedings{zhang2023adding,
  title={Adding conditional control to text-to-image diffusion models},
  author={Zhang, Lvmin and Rao, Anyi and Agrawala, Maneesh},
  booktitle={Proceedings of the IEEE/CVF International Conference on Computer Vision},
  pages={3836--3847},
  year={2023}
}

@inproceedings{fu2022stylegan,
  title={Stylegan-human: A data-centric odyssey of human generation},
  author={Fu, Jianglin and Li, Shikai and Jiang, Yuming and Lin, Kwan-Yee and Qian, Chen and Loy, Chen Change and Wu, Wayne and Liu, Ziwei},
  booktitle={European Conference on Computer Vision},
  pages={1--19},
  year={2022},
  organization={Springer}
}

@inproceedings{reed2016generative,
  title={Generative adversarial text to image synthesis},
  author={Reed, Scott and Akata, Zeynep and Yan, Xinchen and Logeswaran, Lajanugen and Schiele, Bernt and Lee, Honglak},
  booktitle={International conference on machine learning},
  pages={1060--1069},
  year={2016},
  organization={PMLR}
}

@inproceedings{zhang2017stackgan,
  title={Stackgan: Text to photo-realistic image synthesis with stacked generative adversarial networks},
  author={Zhang, Han and Xu, Tao and Li, Hongsheng and Zhang, Shaoting and Wang, Xiaogang and Huang, Xiaolei and Metaxas, Dimitris N},
  booktitle={Proceedings of the IEEE international conference on computer vision},
  pages={5907--5915},
  year={2017}
}

@article{brock2018large,
  title={Large Scale GAN Training for High Fidelity Natural Image Synthesis},
  author={Brock, Andrew},
  journal={arXiv preprint arXiv:1809.11096},
  year={2018}
}

@article{li2019controllable,
  title={Controllable text-to-image generation},
  author={Li, Bowen and Qi, Xiaojuan and Lukasiewicz, Thomas and Torr, Philip},
  journal={Advances in neural information processing systems},
  volume={32},
  year={2019}
}

@InProceedings{Xu_2021_CVPR,
    author    = {Xu, Jianjin and Zheng, Changxi},
    title     = {Linear Semantics in Generative Adversarial Networks},
    booktitle = {Proceedings of the IEEE/CVF Conference on Computer Vision and Pattern Recognition (CVPR)},
    month     = {June},
    year      = {2021},
    pages     = {9351-9360}
}

@article{ho2020denoising,
  title={Denoising diffusion probabilistic models},
  author={Ho, Jonathan and Jain, Ajay and Abbeel, Pieter},
  journal={Advances in neural information processing systems},
  volume={33},
  pages={6840--6851},
  year={2020}
}

@article{song2020denoising,
  title={Denoising diffusion implicit models},
  author={Song, Jiaming and Meng, Chenlin and Ermon, Stefano},
  journal={arXiv preprint arXiv:2010.02502},
  year={2020}
}

@inproceedings{nichol2021improved,
  title={Improved denoising diffusion probabilistic models},
  author={Nichol, Alexander Quinn and Dhariwal, Prafulla},
  booktitle={International conference on machine learning},
  pages={8162--8171},
  year={2021},
  organization={PMLR}
}

@article{podell2023sdxl,
  title={Sdxl: Improving latent diffusion models for high-resolution image synthesis},
  author={Podell, Dustin and English, Zion and Lacey, Kyle and Blattmann, Andreas and Dockhorn, Tim and M{\"u}ller, Jonas and Penna, Joe and Rombach, Robin},
  journal={arXiv preprint arXiv:2307.01952},
  year={2023}
}

@inproceedings{peebles2023scalable,
  title={Scalable diffusion models with transformers},
  author={Peebles, William and Xie, Saining},
  booktitle={Proceedings of the IEEE/CVF International Conference on Computer Vision},
  pages={4195--4205},
  year={2023}
}

@inproceedings{cao2023masactrl,
  title={Masactrl: Tuning-free mutual self-attention control for consistent image synthesis and editing},
  author={Cao, Mingdeng and Wang, Xintao and Qi, Zhongang and Shan, Ying and Qie, Xiaohu and Zheng, Yinqiang},
  booktitle={Proceedings of the IEEE/CVF International Conference on Computer Vision},
  pages={22560--22570},
  year={2023}
}

@article{gal2022image,
  title={An image is worth one word: Personalizing text-to-image generation using textual inversion},
  author={Gal, Rinon and Alaluf, Yuval and Atzmon, Yuval and Patashnik, Or and Bermano, Amit H and Chechik, Gal and Cohen-Or, Daniel},
  journal={arXiv preprint arXiv:2208.01618},
  year={2022}
}

@inproceedings{ruiz2023dreambooth,
  title={Dreambooth: Fine tuning text-to-image diffusion models for subject-driven generation},
  author={Ruiz, Nataniel and Li, Yuanzhen and Jampani, Varun and Pritch, Yael and Rubinstein, Michael and Aberman, Kfir},
  booktitle={Proceedings of the IEEE/CVF conference on computer vision and pattern recognition},
  pages={22500--22510},
  year={2023}
}

@inproceedings{kumari2023multi,
  title={Multi-concept customization of text-to-image diffusion},
  author={Kumari, Nupur and Zhang, Bingliang and Zhang, Richard and Shechtman, Eli and Zhu, Jun-Yan},
  booktitle={Proceedings of the IEEE/CVF Conference on Computer Vision and Pattern Recognition},
  pages={1931--1941},
  year={2023}
}

@article{gu2024mix,
  title={Mix-of-show: Decentralized low-rank adaptation for multi-concept customization of diffusion models},
  author={Gu, Yuchao and Wang, Xintao and Wu, Jay Zhangjie and Shi, Yujun and Chen, Yunpeng and Fan, Zihan and Xiao, Wuyou and Zhao, Rui and Chang, Shuning and Wu, Weijia and others},
  journal={Advances in Neural Information Processing Systems},
  volume={36},
  year={2024}
}

@article{sun2024rectifid,
  title={RectifID: Personalizing Rectified Flow with Anchored Classifier Guidance},
  author={Sun, Zhicheng and Yang, Zhenhao and Jin, Yang and Chi, Haozhe and Xu, Kun and Chen, Liwei and Jiang, Hao and Song, Yang and Gai, Kun and Mu, Yadong},
  journal={arXiv preprint arXiv:2405.14677},
  year={2024}
}

@article{yao2024concept,
  title={Concept Conductor: Orchestrating Multiple Personalized Concepts in Text-to-Image Synthesis},
  author={Yao, Zebin and Feng, Fangxiang and Li, Ruifan and Wang, Xiaojie},
  journal={arXiv preprint arXiv:2408.03632},
  year={2024}
}

@article{wang2024spotactor,
  title={SpotActor: Training-Free Layout-Controlled Consistent Image Generation},
  author={Wang, Jiahao and Yan, Caixia and Zhang, Weizhan and Lin, Haonan and Wang, Mengmeng and Dai, Guang and Gong, Tieliang and Sun, Hao and Wang, Jingdong},
  journal={arXiv preprint arXiv:2409.04801},
  year={2024}
}

@inproceedings{chen2025freecompose,
  title={FreeCompose: Generic Zero-Shot Image Composition with Diffusion Prior},
  author={Chen, Zhekai and Wang, Wen and Yang, Zhen and Yuan, Zeqing and Chen, Hao and Shen, Chunhua},
  booktitle={European Conference on Computer Vision},
  pages={70--87},
  year={2025},
  organization={Springer}
}

@article{yan2023facestudio,
  title={Facestudio: Put your face everywhere in seconds},
  author={Yan, Yuxuan and Zhang, Chi and Wang, Rui and Zhou, Yichao and Zhang, Gege and Cheng, Pei and Yu, Gang and Fu, Bin},
  journal={arXiv preprint arXiv:2312.02663},
  year={2023}
}

@inproceedings{ma2024subject,
  title={Subject-diffusion: Open domain personalized text-to-image generation without test-time fine-tuning},
  author={Ma, Jian and Liang, Junhao and Chen, Chen and Lu, Haonan},
  booktitle={ACM SIGGRAPH 2024 Conference Papers},
  pages={1--12},
  year={2024}
}

@inproceedings{cui2024idadapter,
  title={IDAdapter: Learning Mixed Features for Tuning-Free Personalization of Text-to-Image Models},
  author={Cui, Siying and Guo, Jia and An, Xiang and Deng, Jiankang and Zhao, Yongle and Wei, Xinyu and Feng, Ziyong},
  booktitle={Proceedings of the IEEE/CVF Conference on Computer Vision and Pattern Recognition},
  pages={950--959},
  year={2024}
}

@article{xiao2024fastcomposer,
  title={Fastcomposer: Tuning-free multi-subject image generation with localized attention},
  author={Xiao, Guangxuan and Yin, Tianwei and Freeman, William T and Durand, Fr{\'e}do and Han, Song},
  journal={International Journal of Computer Vision},
  pages={1--20},
  year={2024},
  publisher={Springer}
}

@article{ye2023ip,
  title={Ip-adapter: Text compatible image prompt adapter for text-to-image diffusion models},
  author={Ye, Hu and Zhang, Jun and Liu, Sibo and Han, Xiao and Yang, Wei},
  journal={arXiv preprint arXiv:2308.06721},
  year={2023}
}

@article{wang2024magicface,
  title={MagicFace: Training-free Universal-Style Human Image Customized Synthesis},
  author={Wang, Yibin and Zhang, Weizhong and Jin, Cheng},
  journal={arXiv preprint arXiv:2408.07433},
  year={2024}
}

@article{wang2024taming,
  title={Taming Rectified Flow for Inversion and Editing},
  author={Wang, Jiangshan and Pu, Junfu and Qi, Zhongang and Guo, Jiayi and Ma, Yue and Huang, Nisha and Chen, Yuxin and Li, Xiu and Shan, Ying},
  journal={arXiv preprint arXiv:2411.04746},
  year={2024}
}

@inproceedings{chen2024training,
  title={Training-free layout control with cross-attention guidance},
  author={Chen, Minghao and Laina, Iro and Vedaldi, Andrea},
  booktitle={Proceedings of the IEEE/CVF Winter Conference on Applications of Computer Vision},
  pages={5343--5353},
  year={2024}
}

@inproceedings{liu2025training,
  title={Training-Free Composite Scene Generation for Layout-to-Image Synthesis},
  author={Liu, Jiaqi and Huang, Tao and Xu, Chang},
  booktitle={European Conference on Computer Vision},
  pages={37--53},
  year={2025},
  organization={Springer}
}

@inproceedings{avrahami2023break,
  title={Break-a-scene: Extracting multiple concepts from a single image},
  author={Avrahami, Omri and Aberman, Kfir and Fried, Ohad and Cohen-Or, Daniel and Lischinski, Dani},
  booktitle={SIGGRAPH Asia 2023 Conference Papers},
  pages={1--12},
  year={2023}
}

@article{jang2024identity,
  title={Identity Decoupling for Multi-Subject Personalization of Text-to-Image Models},
  author={Jang, Sangwon and Jo, Jaehyeong and Lee, Kimin and Hwang, Sung Ju},
  journal={arXiv preprint arXiv:2404.04243},
  year={2024}
}

@article{kim2024instantfamily,
  title={InstantFamily: Masked Attention for Zero-shot Multi-ID Image Generation},
  author={Kim, Chanran and Lee, Jeongin and Joung, Shichang and Kim, Bongmo and Baek, Yeul-Min},
  journal={arXiv preprint arXiv:2404.19427},
  year={2024}
}

@inproceedings{kong2025omg,
  title={Omg: Occlusion-friendly personalized multi-concept generation in diffusion models},
  author={Kong, Zhe and Zhang, Yong and Yang, Tianyu and Wang, Tao and Zhang, Kaihao and Wu, Bizhu and Chen, Guanying and Liu, Wei and Luo, Wenhan},
  booktitle={European Conference on Computer Vision},
  pages={253--270},
  year={2025},
  organization={Springer}
}

@inproceedings{kwon2024concept,
  title={Concept Weaver: Enabling Multi-Concept Fusion in Text-to-Image Models},
  author={Kwon, Gihyun and Jenni, Simon and Li, Dingzeyu and Lee, Joon-Young and Ye, Jong Chul and Heilbron, Fabian Caba},
  booktitle={Proceedings of the IEEE/CVF Conference on Computer Vision and Pattern Recognition},
  pages={8880--8889},
  year={2024}
}

@inproceedings{zheng2023layoutdiffusion,
  title={Layoutdiffusion: Controllable diffusion model for layout-to-image generation},
  author={Zheng, Guangcong and Zhou, Xianpan and Li, Xuewei and Qi, Zhongang and Shan, Ying and Li, Xi},
  booktitle={Proceedings of the IEEE/CVF Conference on Computer Vision and Pattern Recognition},
  pages={22490--22499},
  year={2023}
}

@inproceedings{sarukkai2024collage,
  title={Collage diffusion},
  author={Sarukkai, Vishnu and Li, Linden and Ma, Arden and R{\'e}, Christopher and Fatahalian, Kayvon},
  booktitle={Proceedings of the IEEE/CVF Winter Conference on Applications of Computer Vision},
  pages={4208--4217},
  year={2024}
}

@article{wang2024ms,
  title={MS-Diffusion: Multi-subject Zero-shot Image Personalization with Layout Guidance},
  author={Wang, X and Fu, Siming and Huang, Qihan and He, Wanggui and Jiang, Hao},
  journal={arXiv preprint arXiv:2406.07209},
  year={2024}
}

@inproceedings{zhou2024migc,
  title={Migc: Multi-instance generation controller for text-to-image synthesis},
  author={Zhou, Dewei and Li, You and Ma, Fan and Zhang, Xiaoting and Yang, Yi},
  booktitle={Proceedings of the IEEE/CVF Conference on Computer Vision and Pattern Recognition},
  pages={6818--6828},
  year={2024}
}

@inproceedings{hoe2024interactdiffusion,
  title={InteractDiffusion: Interaction Control in Text-to-Image Diffusion Models},
  author={Hoe, Jiun Tian and Jiang, Xudong and Chan, Chee Seng and Tan, Yap-Peng and Hu, Weipeng},
  booktitle={Proceedings of the IEEE/CVF Conference on Computer Vision and Pattern Recognition},
  pages={6180--6189},
  year={2024}
}

@inproceedings{wang2024moa,
  title={Moa: Mixture-of-attention for subject-context disentanglement in personalized image generation},
  author={Wang, Kuan-Chieh and Ostashev, Daniil and Fang, Yuwei and Tulyakov, Sergey and Aberman, Kfir},
  booktitle={SIGGRAPH Asia 2024 Conference Papers},
  pages={1--12},
  year={2024}
}

@article{achiam2023gpt,
  title={Gpt-4 technical report},
  author={Achiam, Josh and Adler, Steven and Agarwal, Sandhini and Ahmad, Lama and Akkaya, Ilge and Aleman, Florencia Leoni and Almeida, Diogo and Altenschmidt, Janko and Altman, Sam and Anadkat, Shyamal and others},
  journal={arXiv preprint arXiv:2303.08774},
  year={2023}
}

@article{radford2018improving,
  title={Improving language understanding by generative pre-training},
  author={Radford, Alec},
  year={2018}
}

@misc{liu2023improvedllava,
      title={Improved Baselines with Visual Instruction Tuning}, 
      author={Liu, Haotian and Li, Chunyuan and Li, Yuheng and Lee, Yong Jae},
      publisher={arXiv:2310.03744},
      year={2023},
}

@misc{liu2023llava,
      title={Visual Instruction Tuning}, 
      author={Liu, Haotian and Li, Chunyuan and Wu, Qingyang and Lee, Yong Jae},
      publisher={NeurIPS},
      year={2023},
}

@article{liu2023grounding,
  title={Grounding dino: Marrying dino with grounded pre-training for open-set object detection},
  author={Liu, Shilong and Zeng, Zhaoyang and Ren, Tianhe and Li, Feng and Zhang, Hao and Yang, Jie and Li, Chunyuan and Yang, Jianwei and Su, Hang and Zhu, Jun and others},
  journal={arXiv preprint arXiv:2303.05499},
  year={2023}
}

@inproceedings{caron2021emerging,
  title={Emerging Properties in Self-Supervised Vision Transformers},
  author={Caron, Mathilde and Touvron, Hugo and Misra, Ishan and J\'egou, Herv\'e  and Mairal, Julien and Bojanowski, Piotr and Joulin, Armand},
  booktitle={Proceedings of the International Conference on Computer Vision (ICCV)},
  year={2021}
}

@inproceedings{wang2024instancediffusion,
  title={Instancediffusion: Instance-level control for image generation},
  author={Wang, Xudong and Darrell, Trevor and Rambhatla, Sai Saketh and Girdhar, Rohit and Misra, Ishan},
  booktitle={Proceedings of the IEEE/CVF Conference on Computer Vision and Pattern Recognition},
  pages={6232--6242},
  year={2024}
}

@article{wu2023human,
  title={Human Preference Score v2: A Solid Benchmark for Evaluating Human Preferences of Text-to-Image Synthesis},
  author={Wu, Xiaoshi and Hao, Yiming and Sun, Keqiang and Chen, Yixiong and Zhu, Feng and Zhao, Rui and Li, Hongsheng},
  journal={arXiv preprint arXiv:2306.09341},
  year={2023}
}

@inproceedings{radford2021learning,
  title={Learning transferable visual models from natural language supervision},
  author={Radford, Alec and Kim, Jong Wook and Hallacy, Chris and Ramesh, Aditya and Goh, Gabriel and Agarwal, Sandhini and Sastry, Girish and Askell, Amanda and Mishkin, Pamela and Clark, Jack and others},
  booktitle={International conference on machine learning},
  pages={8748--8763},
  year={2021},
  organization={PMLR}
}

@inproceedings{Deng2020CVPR,
title = {RetinaFace: Single-Shot Multi-Level Face Localisation in the Wild},
author = {Deng, Jiankang and Guo, Jia and Ververas, Evangelos and Kotsia, Irene and Zafeiriou, Stefanos},
booktitle = {CVPR},
year = {2020}
}

@inproceedings{schroff2015facenet,
  title={Facenet: A unified embedding for face recognition and clustering},
  author={Schroff, Florian and Kalenichenko, Dmitry and Philbin, James},
  booktitle={Proceedings of the IEEE conference on computer vision and pattern recognition},
  pages={815--823},
  year={2015}
}

@inproceedings{ronneberger2015u,
  title={U-net: Convolutional networks for biomedical image segmentation},
  author={Ronneberger, Olaf and Fischer, Philipp and Brox, Thomas},
  booktitle={Medical image computing and computer-assisted intervention--MICCAI 2015: 18th international conference, Munich, Germany, October 5-9, 2015, proceedings, part III 18},
  pages={234--241},
  year={2015},
  organization={Springer}
}

@Misc{omost,
  author = {Omost Team},
  title  = {Omost GitHub Page},
  year   = {2024},
}

@inproceedings{brooks2023instructpix2pix,
  title={Instructpix2pix: Learning to follow image editing instructions},
  author={Brooks, Tim and Holynski, Aleksander and Efros, Alexei A},
  booktitle={Proceedings of the IEEE/CVF conference on computer vision and pattern recognition},
  pages={18392--18402},
  year={2023}
}

@inproceedings{kawar2023imagic,
  title={Imagic: Text-based real image editing with diffusion models},
  author={Kawar, Bahjat and Zada, Shiran and Lang, Oran and Tov, Omer and Chang, Huiwen and Dekel, Tali and Mosseri, Inbar and Irani, Michal},
  booktitle={Proceedings of the IEEE/CVF conference on computer vision and pattern recognition},
  pages={6007--6017},
  year={2023}
}

@article{poole2022dreamfusion,
  title={Dreamfusion: Text-to-3d using 2d diffusion},
  author={Poole, Ben and Jain, Ajay and Barron, Jonathan T and Mildenhall, Ben},
  journal={arXiv preprint arXiv:2209.14988},
  year={2022}
}

@inproceedings{chen2023text2tex,
  title={Text2tex: Text-driven texture synthesis via diffusion models},
  author={Chen, Dave Zhenyu and Siddiqui, Yawar and Lee, Hsin-Ying and Tulyakov, Sergey and Nie{\ss}ner, Matthias},
  booktitle={Proceedings of the IEEE/CVF international conference on computer vision},
  pages={18558--18568},
  year={2023}
}

@misc{lugmayr2022repaintinpaintingusingdenoising,
      title={RePaint: Inpainting using Denoising Diffusion Probabilistic Models}, 
      author={Andreas Lugmayr and Martin Danelljan and Andres Romero and Fisher Yu and Radu Timofte and Luc Van Gool},
      year={2022},
      eprint={2201.09865},
      archivePrefix={arXiv},
      primaryClass={cs.CV},
      url={https://arxiv.org/abs/2201.09865}, 
}
}
\clearpage
\setcounter{page}{1}
\maketitlesupplementary

\renewcommand\thesection{S\arabic{section}}
\renewcommand\thesubsection{S\arabic{section}.\arabic{subsection}}
\renewcommand{\thetable}{S\arabic{table}}
\renewcommand{\thefigure}{S\arabic{figure}}

\setcounter{section}{0}
\setcounter{table}{0}
\setcounter{figure}{0}
\setcounter{equation}{0}

\section{\benchmark{}}
\label{sec:idbench}

\begin{figure*}
    \centering
    \includegraphics[width=0.99\linewidth]{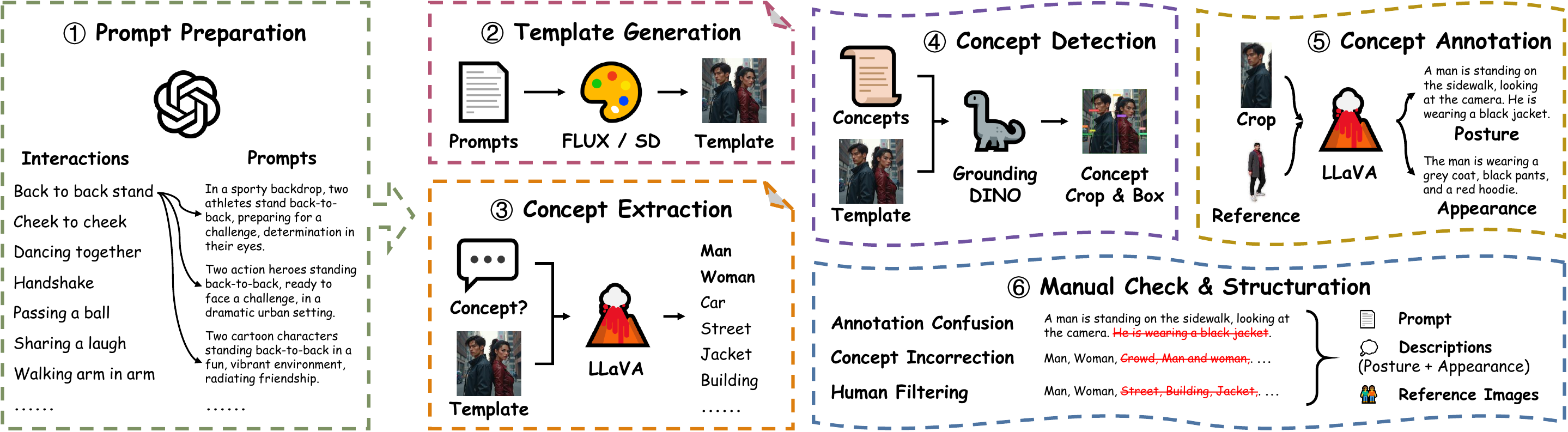}
    \caption{
    The construction pipeline of \benchmark{}.
    }
    \label{fig:idbench}
\end{figure*}

Figure \ref{fig:idbench} illustrates the construction pipeline of \benchmark{}.
First, a significant characteristic of multi-ID images is the conspicuous interaction between different identities.
The benchmark construction starts with nominated interaction generation.
We leverage LLM \cite{radford2018improving,achiam2023gpt} as an advisor to provide phrases about common interactions and generate detailed prompts based on them.
Then, with the advantage of text-to-image generation models \cite{esser2024scaling,flux2023}, we transform texts into intricate visual templates, thus achieving enhanced posture and appearance representations.
We next capture dominated concepts from prompts and template images with LLaVA \cite{liu2023llava,liu2023improvedllava} and detect instances of these concepts using Grounding DINO \cite{caron2021emerging,liu2023grounding}.
Subsequently, we utilize LLaVA again to obtain posture and appearance descriptions from cropped concepts and reference images, respectively.
The reference images adopted as personalization ID are mainly collected from an open-source human image dataset \cite{fu2022stylegan}. 
Finally, we manually retain human concepts and correct errors such as confused annotations, inappropriate concepts, etc., acquiring structured samples. 

{\small\Circled{1}} Prompt Preparation: 
We use the prompt of ``\textit{Please help me generate interactions between two people, such as `Back-to-back stand'}'' to generate 40 different phrases about multi-ID interaction.
Then, we generate 10 prompts for each interaction using prompts like ``\textit{Please generate 10 prompts about two `Back-to-back stand' people, which will be used as conditions for text-to-image generation}''.

{\small\Circled{2}} Template Generation: 
We add a prefix of ``\textit{A portrait of 2 people}'' before generated prompts, and use these prompts as conditions of \texttt{FLUX.1-dev} \cite{flux2023} to generate template images.

{\small\Circled{3}} Concept Extraction: 
The text condition of LLaVA model is set to 
``\textit{Please list the types of objects or concepts in this image. Each concept only needs to be listed once, and the essential components in the image should be listed as much as possible. Examine the provided image and identify the main components within it. Please extract a list of relevant nouns, ensuring the focus is on people and the most significant elements present in the image. Aim to identify no more than ten objects or individuals. Ensure that the selected nouns accurately represent the key components, such as: Individuals (e.g., 'man', 'woman', 'child'), Main objects (e.g., 'car', 'tree', 'building'). Return the list in a concise format. Focus on significant elements like people (without details on clothing, accessories, or expressions), and main objects or concepts of the environment. Exclude any minor details that are not central to the composition. Please separate each concept with a comma, e.g.: "table, man, apple". Try not to exceed ten concepts.}''
After obtaining accurate image components, we use sentence extension with conversational questioning to obtain the concept of ID: 
``\textit{Question: "Does this word "man" correspond to a category of people?" Answer: Yes. Question: "Does this word "\%s" correspond to a category of people?" Answer: }''.

{\small\Circled{4}} Concept Detection:
With the help of Grounding DINO, we detect the concepts about ID generated in the previous stage. 
We store the bounding boxes and crop the corresponding concepts.

{\small\Circled{5}} Concept Annotation:
We first annotate the appearance of our collected reference images using LLaVA with the prompt of 
``\textit{
This is an image of a person. Please describe this image in detail. Generate detailed descriptions, organize your observations into two distinct sections: 'State' and 'Appearance'. 'State': Describe the actions, expressions, and poses of any individuals in the image, as well as the positions, angles, and conditions of objects. 'Consider aspects such as: Actions or motions (e.g., 'running', 'sitting'), Facial expressions (e.g., 'smiling', 'frowning'), Postures (e.g., 'standing upright', 'slouched'), Object status (e.g., 'broken', 'new') and orientation (e.g., 'tilted', 'upright'). 'Appearance': Detail the physical characteristics of individuals and objects, including: Human features (e.g., 'hair color', 'gender', 'age'), Clothing details (e.g., 'color', 'style', 'fit'), Object characteristics (e.g., 'color', 'texture', 'size'). Provide a comprehensive description with two parts of the 'State' and 'Appearance' of the mentioned concept in the image. Both parts should not be longer than 50 words. Please provide only the descriptions directly, the description should be as detailed as possible
}''.
Then we retain the part after ``\textit{Appearance}'' as appearance descriptions of reference images.
Subsequently, we use this prompt to generate descriptions of human concepts cropped from template images.
On the contrary, we keep the ``'\textit{State}' part as posture descriptions of our benchmark samples.

{\small\Circled{6}} Manual Check \& Structuration:
To ensure the accuracy of benchmark samples, we manually checked the accuracy of the generated description and the correctness of the components (whether only posture or appearance is described).
Subsequently, we randomly assign a high-level semantically consistent (man or woman, adult or child, etc.) reference image to each human concept in generated samples as ID conditions.
The text condition of each ID is composed of a combination of posture description (from the template crop) and appearance description (from the reference image).
Each final benchmark sample contains a global prompt, high-level semantic category labels for multiple IDs, ID reference images, ID posture descriptions, and ID full text descriptions.

\section{More Multi-ID Personalization Results}

\begin{figure*}[h]
    \centering
    \includegraphics[width=0.99\linewidth]{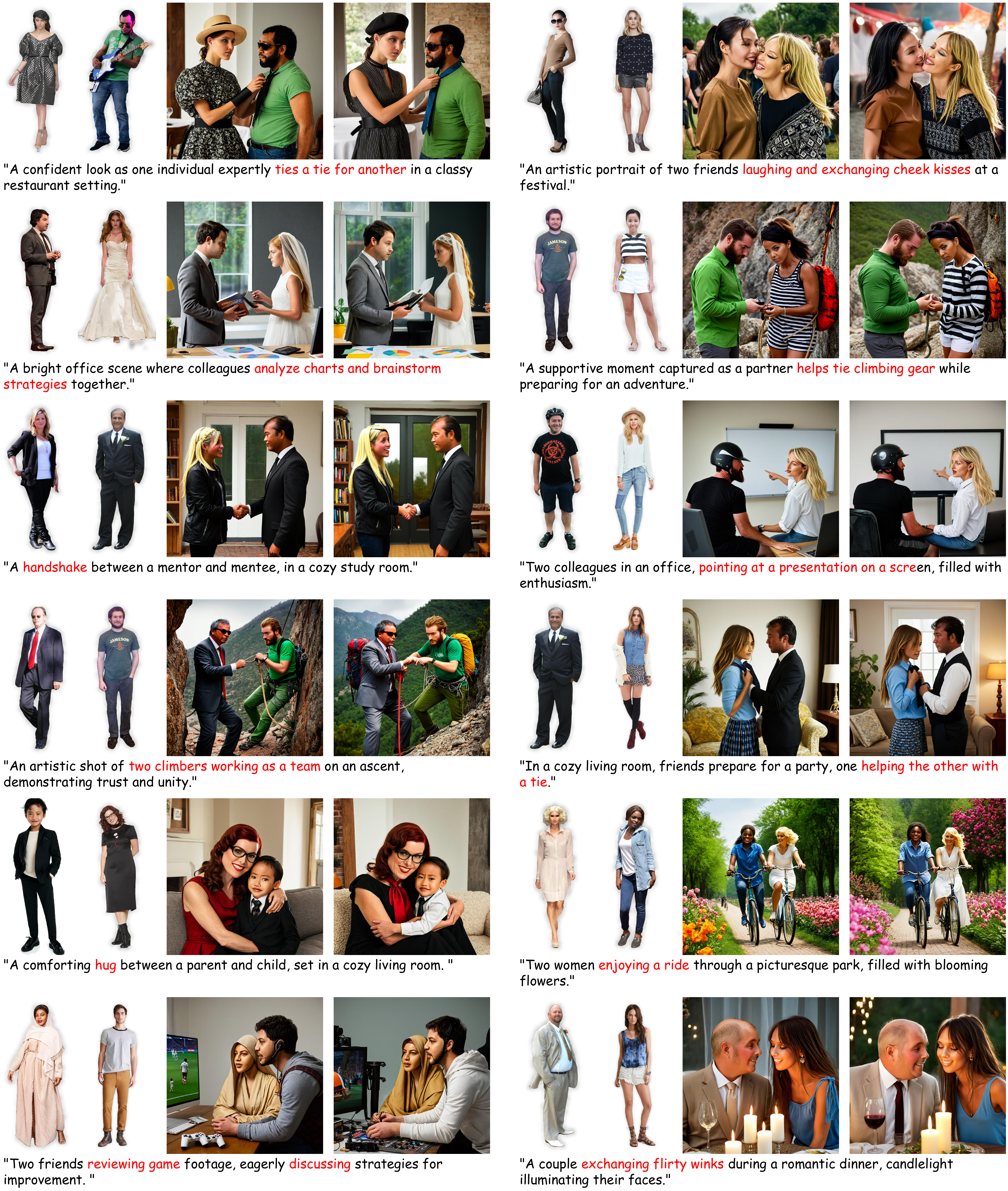}
    \caption{
    Detailed qualitative results of \approach{} on \benchmark{}.
    We {\color{red} highlight} the description of human interactions in prompts, emphasizing the comparison of human postures in the generated images.
    }
    \label{fig:supp_vis}
\end{figure*}

In this section, we showcase additional detailed examples generated using our proposed \approach{} in Fig.\ref{fig:supp_vis} to illustrate the effectiveness of our approach.

\end{document}